\documentclass{article}
\usepackage{graphicx} 
\usepackage{biblatex}
\usepackage{booktabs}
\usepackage{tabularx}  
\usepackage{subcaption}
\usepackage{authblk} 

\title{Optimizing the Interface  Between \\ Knowledge Graphs and LLMs \\ for Complex Reasoning\thanks{This is a preliminary version. A revised and expanded version is in preparation.} }
\author[1]{Vasilije Markovi\' c}
\author[1]{Lazar Obradovi\' c}
\author[1,2,3]{ Laszlo Hajdu}
\author[3]{Jovan Pavlovi\' c}

\affil[1]{Cognee Inc.}
\affil[2]{Innorenew CoE}
\affil[3]{University of Primorska, FAMNIT}

\date{}

\begin{document}

\maketitle

\begin{abstract}
Integrating Large Language Models (LLMs) with Knowledge Graphs (KGs) results in complex systems with numerous hyperparameters that directly affect performance. While such systems are increasingly common in retrieval-augmented generation, the role of systematic hyperparameter optimization remains underexplored. In this paper, we study this problem in the context of Cognee, a modular framework for end-to-end KG construction and retrieval. Using three multi-hop QA benchmarks (HotPotQA, TwoWikiMultiHop, and MuSiQue) we optimize parameters related to chunking, graph construction, retrieval, and prompting. Each configuration is scored using established metrics (exact match, F1, and DeepEval’s LLM-based \emph{correctness} metric). Our results demonstrate that meaningful gains can be achieved through targeted tuning. While the gains are consistent, they are not uniform, with performance varying across datasets and metrics. This variability highlights both the value of tuning and the limitations of standard evaluation measures. While demonstrating the immediate potential of hyperparameter tuning, we argue that future progress will depend not only on architectural advances but also on clearer frameworks for optimization and evaluation in complex, modular systems.
\end{abstract}
\section{Introduction}

Large Language Models (LLMs), based on Transformer architectures \cite{vaswani2017attention}, have demonstrated strong performance across a wide range of natural language processing tasks, including open-domain question answering, summarization, and generation. Their impact has been felt across diverse application areas, from healthcare and finance to legal and scientific domains \cite{medical_app_2, fin_app_1, legal_app_2, sci-disc_app}. While these models store vast amounts of information in their parameters \cite{knowledge1, knowledge2, trust}, they are also prone to producing confident but incorrect outputs \cite{hall1, hall2}. In addition, they lack an efficient mechanism for updating or extending knowledge without retraining.

Retrieval-Augmented Generation (RAG) has emerged as a standard approach to mitigate these issues \cite{rag_survey_1}. In typical RAG pipelines, a dense retriever selects relevant textual context for a given query, and the retrieved content is appended to the query before being processed by the LLM. This design improves factuality and allows models to reference external sources. However, standard RAG systems often struggle with questions that involve multi-step reasoning or require structured access to relational knowledge. In such cases, relying solely on dense or sparse document retrieval is insufficient \cite{shared_keyword, microsoft_local}.

To address these limitations, hybrid approaches that integrate knowledge graphs into RAG workflows have gained attention. These systems, sometimes referred to as GraphRAG, use graphs to represent relational structures and support retrieval based on symbolic queries or multi-hop graph traversal \cite{survey_1, survey_2}. Graph-based retrieval augments LLMs with access to explicit, structured context and has shown promise in tasks requiring deeper reasoning.

One challenge that persists across both classical and graph-based RAG systems is hyperparameter sensitivity. The performance of these pipelines depends heavily on a wide range of configuration choices, including chunk size, retriever type, top-$k$ thresholds, and prompt templates. As pipelines grow more modular and sophisticated, the number of tunable parameters increases, and their interactions become more complex. While hyperparameter optimization has been explored in standard RAG systems, its role in graph-enhanced pipelines remains underexamined.

This paper addresses that gap. We present a structured study of hyperparameter optimization in graph-based RAG systems, with a focus on tasks that combine unstructured inputs, knowledge graph construction, retrieval, and generation. Our experiments use Cognee, an open-source framework that supports end-to-end graph-based memory construction and retrieval. Cognee’s modularity allows for clean separation and independent configuration of pipeline components, making it well-suited for controlled optimization studies.

We evaluate on three established multi-hop QA benchmarks: HotPotQA, TwoWikiMultiHop, and Musique. Each configuration is scored using one of three metrics: exact match (EM), token-level F1, or correctness. The correctness score is computed using DeepEval, an LLM-based grading tool that evaluates answer plausibility against the gold reference. Our study demonstrates that even modest parameter changes can lead to measurable improvements, but also that metric choice and task characteristics strongly influence outcomes.

The next section reviews related work on retrieval-augmented generation, graph-based systems, and hyperparameter optimization. In Section~\ref{sec:cognee}, we describe the Cognee framework and its architecture for knowledge graph construction and retrieval. Section~\ref{sec:optimization} outlines the hyperparameter optimization setup, including the parameter space and optimization method. Section~\ref{sec:experimental-setup} details the experimental design, benchmarks, and evaluation metrics. Section~\ref{sec:results} presents the results and discusses their implications. We conclude with a summary of findings and directions for future work in Section~\ref{sec:conclusion}.
\section{Background and Related Work}
We review key developments relevant to our work, focusing on retrieval-augmented generation (RAG), multi-hop and graph-based question answering, and recent advances in hyperparameter optimization for LLM pipelines. Particular attention is given to methods that combine structured retrieval with neural generation and those that treat pipeline configuration as an optimization problem.

\paragraph{Advances and Challenges in RAG Systems}

Retrieval-augmented generation (RAG) systems extend language models with a retrieval module to ground outputs in external knowledge \cite{rag_pioneer1, rag_pioneer2}. This basic two-stage architecture has become the de facto standard, with numerous refinements proposed over time \cite{rag_survey_1}. Recent work includes Self-RAG \cite{selfrag}, which enables LLMs to reflect on their outputs and dynamically trigger retrieval, and CRAG \cite{crag}, which filters low-confidence documents using a retrieval evaluator and escalates to web search when needed. Comprehensive surveys summarize the state of the field and its variants \cite{rag_survey_2, rag_survey_3, rag_survey_4}.

\paragraph{Multi-Hop Question Answering}

Multi-hop QA extends standard QA by requiring reasoning over multiple documents. Early datasets like HotPotQA \cite{hotpotqa} crowdsourced such questions over Wikipedia. 2WikiMultiHopQA \cite{2wiki} improves on this by leveraging Wikidata relations to enforce structured, verifiable reasoning paths. MuSiQue \cite{musique} takes a bottom-up approach, composing multi-step questions from single-hop primitives and filtering out spurious shortcuts, offering a more robust benchmark for compositional reasoning.

\paragraph{Knowledge Graph Question Answering}

KGQA systems answer questions via structured reasoning over graphs \cite{kgqa1, kgqa2, kgqa3, kgqa4}, increasingly integrating LLMs to bridge symbolic and neural reasoning \cite{kg_roadmap, kg_survey_2}. RoG \cite{luo2023reasoning} prompts LLMs to generate abstract relation paths that are grounded via graph traversal before final answer generation. Other work includes trainable subgraph retrievers \cite{subgraph_retriever} and decomposed logical reasoning chains over subgraphs \cite{choudhary2023complex}, demonstrating measurable gains in both interpretability and performance.

\paragraph{GraphRAG}

GraphRAG generalizes RAG to arbitrary graph structures, extending its use beyond knowledge bases \cite{survey_1, survey_2}. Early systems like Microsoft’s summarization pipeline \cite{microsoft_local} use LLMs to build knowledge graphs, partition them with community detection \cite{traag2019louvain}, and summarize each component. Other variants use GNNs with subgraph selection \cite{g-retriever}, graph-traversal agents \cite{shared_keyword}, or Personalized PageRank over schemaless graphs \cite{hipporag}. These systems span a wide range of tasks but share a common structure: dynamic subgraph construction followed by prompt-based reasoning.

\paragraph{Hyperparameter Optimization in RAG}

Optimizing RAG systems requires balancing retrieval coverage, generation accuracy, and resource constraints. Recent work applies Bayesian optimization under budget limits \cite{wang2023cost}, formulates context usage as a tunable variable \cite{cwu}, and introduces full-pipeline tuning via reinforcement learning \cite{multiagent, fu2024autorag}. Multi-objective frameworks have also emerged to trade off accuracy, latency, and safety \cite{multiobjective}. While methodologically diverse, all aim to expose and control the critical degrees of freedom in modern RAG pipelines.
\section{Automated Knowledge Graph Construction: Cognee}
\label{sec:cognee}

Cognee is an open-source framework for end-to-end knowledge graph (KG) construction, retrieval, and completion. It supports heterogeneous inputs (e.g., text, images, audio) from which it extracts entities and relations, possibly with the support of an ontology schema. The extraction process runs in containerized environments and is based on tasks and pipelines, with each stage extensible via configuration or code.

The default pipeline includes ingestion, chunking, large language model (LLM)-based extraction, and indexing into graph, relational, and vector store backends. Downstream of indexing, Cognee provides built-in components for retrieval and completion. A unified interface supports vector search, symbolic graph queries, and hybrid graph–text methods. Completion builds on the same infrastructure, enabling both prompt-based LLM interaction and structured graph queries. 

Cognee also includes a configurable evaluation framework for benchmarking retrieval and completion workflows. The framework is based on multi-hop question answering, providing a structured evaluation setting for graph-based systems using established benchmarks (HotPotQA, TwoWikiMultiHop). Evaluation proceeds sequentially through distinct phases. It begins with corpus construction, followed by context-conditioned answering that leverages the retrieval and completion components. Answers are then compared to gold references and graded using multiple metrics. The final output includes confidence-scored performance reports. For more technical details, including the ingestion, retrieval, and evaluation architecture, see Appendix~\ref{appendix:cognee}.

Cognees’s modularity enables targeted hyperparameter tuning across ingestion, retrieval, and completion stages. The evaluation framework provides structured, quantitative feedback, allowing the system as a whole to be treated as an objective function. This setup enables the direct application of standard hyperparameter optimization algorithms, which we describe in the following section.
\section{Hyperparameter Optimization Setup}
\label{sec:optimization}

This section describes the structure of the optimization process and the parameters explored during tuning. We first outline the end-to-end setup and optimization method, followed by a detailed description of the tunable parameters.

\subsection{Optimization Framework}

\begin{table}[t]
\centering
\begin{tabularx}{\linewidth}{@{}lX@{}}
\toprule
\textbf{Parameter} & \textbf{Description} \\
\midrule
Chunk size & Number of tokens per document segment used during graph extraction \\
Retriever type & Strategy used to retrieve context (text chunks or graph triplets) \\
Top-$k$ & Number of retrieved items passed to the language model \\
QA prompt & Instruction template used for answer generation \\
Graph prompt & Template guiding entity and relation extraction during graph construction \\
Task getter & Configuration for dataset preprocessing and summary handling \\
\bottomrule
\end{tabularx}
\caption{Parameters used during Dreamify optimization.}
\label{tab:dreamify-params}
\end{table}

Cognee exposes multiple configurable components that influence retrieval and generation behavior. These include parameters related to preprocessing, retriever selection, prompt design, and runtime settings. While default values are often selected heuristically, their impact on performance is not always predictable. To evaluate the effect of these design choices systematically, we developed a hyperparameter optimization framework named Dreamify.

Dreamify treats the entire Cognee pipeline as a parameterized process. This includes ingestion, chunking, LLM-based extraction, retrieval, and evaluation. A single configuration defines the behavior of all stages. Each trial corresponds to a complete pipeline run, starting from corpus construction and ending with evaluation against a benchmark dataset. The output is a scalar score based on one of several metrics, such as F1, exact match, or LLM-based correctness. These metrics are computed as averages over all questions in the dataset and return values between 0 and 1.

The optimization is performed using a Tree-structured Parzen Estimator (TPE). This algorithm is well suited to the search space in question, which combines categorical and ordered integer-valued parameters. Grid search is not practical at this scale, and random search underperformed in early tests. While TPE was sufficient for our experiments, other optimization strategies remain open for future work.

The pipeline behavior is deterministic with respect to a fixed configuration, though some components, such as LLM-generated graph construction, exhibit minor variation across runs. These differences do not materially affect overall evaluation scores within a single configuration. Trials are independent and reproducible.  Further experimental details and results are presented in Section~\ref{sec:experimental-setup}.

\subsection{Tunable Parameters}

The optimization process considers six core parameters that influence document processing, retrieval behavior, prompt selection, and graph construction. Each parameter affects how information is segmented, retrieved, or used during answer generation. Table~\ref{tab:dreamify-params} provides a summary; below we describe each parameter in more detail.

\paragraph{Chunk Size (\texttt{chunk\_size})}
This parameter controls the number of tokens used to segment documents before graph extraction. In the Cognee pipeline, it influences both the structure of the resulting graph and the granularity of context available during retrieval. The range used in this study (200–2000 tokens) was chosen based on preliminary testing to balance extraction accuracy, retrieval specificity, and processing time.

\paragraph{Retrieval Strategy (\texttt{search\_type})}
This parameter determines how context is selected for answer generation. The \texttt{cognee\_completion} strategy retrieves text chunks using vector search and passes them directly to the language model. The \texttt{cognee\_graph\_completion} strategy retrieves knowledge graph nodes and their associated triplets by combining vector similarity with graph structure. Retrieved nodes are briefly described, and the surrounding triplets are formatted as structured text. The structured format of retrieved nodes and triplets emphasizes relational context and may support more effective multi-hop reasoning.

\paragraph{Top-K Context Size (\texttt{top\_k})}
This parameter sets the number of items retrieved per query. With \texttt{cognee\_completion}, it controls the number of text chunks; with \texttt{cognee\_graph\_completion}, it controls the number of graph triplets. The retrieved context is passed to the language model for answer generation. In our experiments, values ranged from 1 to 20.

\paragraph{QA Prompt Template (\texttt{qa\_system\_prompt})}
This parameter selects an instruction template used for answer generation. Templates differ in style and specificity, ranging from concise prompts to more detailed instructions that encourage justification or structured outputs. Prompt selection can influence both answer format and factual precision.

\paragraph{Prompt Templates (\texttt{qa\_system\_prompt}, \texttt{graph\_prompt})}
These parameters control the instruction templates used during answer generation and graph construction. For question answering, we evaluated three prompt variants differing primarily in tone and verbosity. While the underlying instruction remained consistent, more constrained and direct prompts often produced outputs that aligned more closely with the expected answer format. This had a notable impact on evaluation scores, particularly for exact match and F1, though correctness scores were also affected to a lesser degree. For graph construction, three prompts were also tested, differing in how they guided the LLM to extract entities and relations from text—either in a single step or through more structured, incremental instructions. This choice influenced the granularity and consistency of the resulting graph structures used during retrieval.

\paragraph{Task Processing Method (\texttt{task\_getter\_type})}
This parameter controls how question–answer pairs are preprocessed during evaluation. While the system can support arbitrary pipeline variants, we focus on two representative configurations. In the first, document summaries are generated during graph construction and made available to the retriever. In the second, summary generation is omitted. 

\section{Experimental Setup}
\label{sec:experimental-setup}

We conducted a series of nine hyperparameter optimization experiments to evaluate the impact of configuration choices on Cognee’s end-to-end performance. Each experiment corresponds to a distinct combination of benchmark dataset and evaluation metric. The datasets used were HotPotQA, TwoWikiMultiHop, and Musique. Each experiment targeted one of three metrics: exact match (EM), F1, or DeepEval’s LLM-based correctness.

For each experiment, we created a filtered subset of the benchmark. Instances were randomly sampled and then manually reviewed prior to experimentation. We excluded examples that were ungrammatical, ambiguous, mislabeled, or unsupported by the provided context. Similar issues have been noted in prior literature. The resulting evaluation set consisted of 24 training instances and 12 test instances per dataset. This filtering step was performed once, before any tuning, to avoid bias or cherry picking.

Within each trial, the knowledge graph was constructed using all context passages from the training set. This produced a single merged graph per trial, which was then used to answer all training questions. The structure of the pipeline remained consistent across all datasets and metrics.

Each experiment consisted of 50 trials. In each trial, a configuration was sampled by the optimizer and executed as a full pipeline run, including ingestion, graph construction, retrieval, and answer generation. The selected metric was computed over all training questions, and the resulting score was used as the objective value for the trial. EM and F1 scores were computed deterministically. The DeepEval correctness score required a separate LLM-based evaluation step.

Trials were run sequentially without parallelization. Execution time per trial was approximately 30 minutes. Final results report performance on the test set using the best-performing configuration selected from training. In addition to point estimates, we report confidence intervals computed using non-parametric bootstrap resampling over individual QA pairs.

\section{Results and Discussion}
\label{sec:results}

In this section, we summarize the outcomes of the optimization experiments, including training and hold-out set performance. We also discuss generalization, parameter effects, and broader implications in the context of knowledge-graph-based retrieval-augmented systems.

\subsection{Training Set Performance}

\begin{figure}[h]
\centering
\begin{subfigure}[t]{\linewidth}
    \centering
    \includegraphics[width=0.9\linewidth]{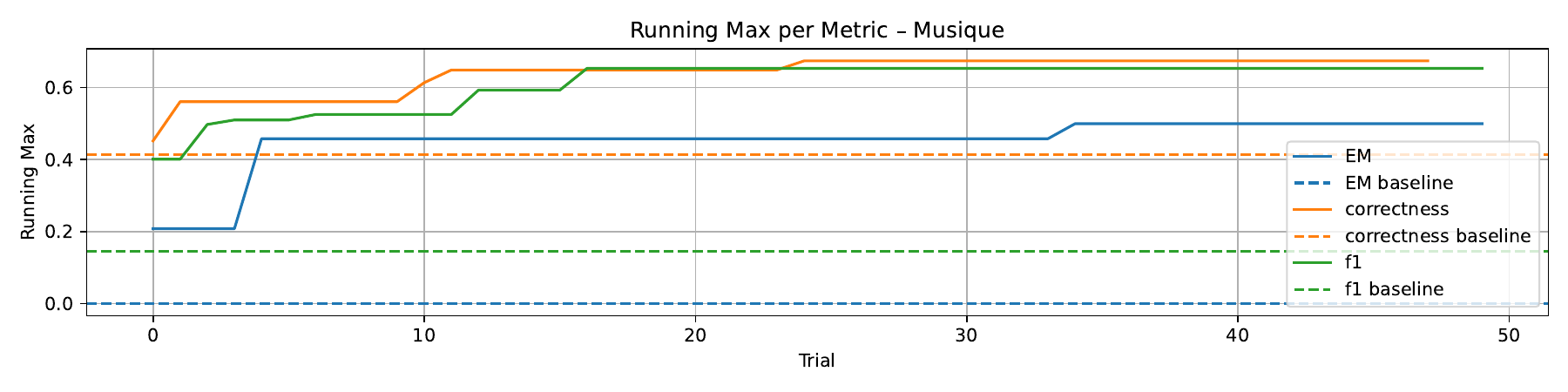}
    \caption{Musique}
    \label{fig:musique}
\end{subfigure}

\vspace{0.5em}

\begin{subfigure}[t]{\linewidth}
    \centering
    \includegraphics[width=0.9\linewidth]{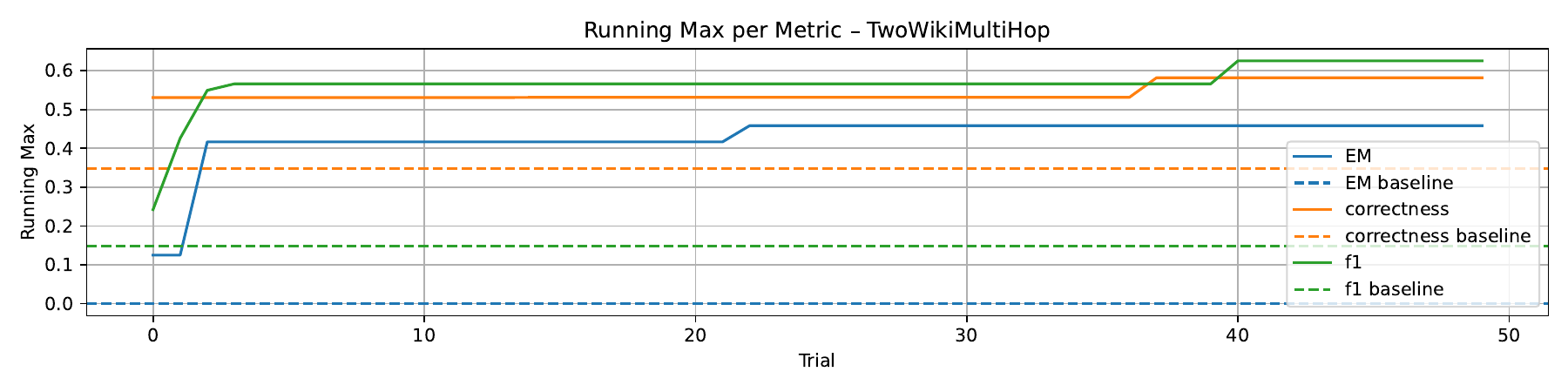}
    \caption{TwoWikiMultiHop}
    \label{fig:twowiki}
\end{subfigure}

\vspace{0.5em}

\begin{subfigure}[t]{\linewidth}
    \centering
    \includegraphics[width=0.9\linewidth]{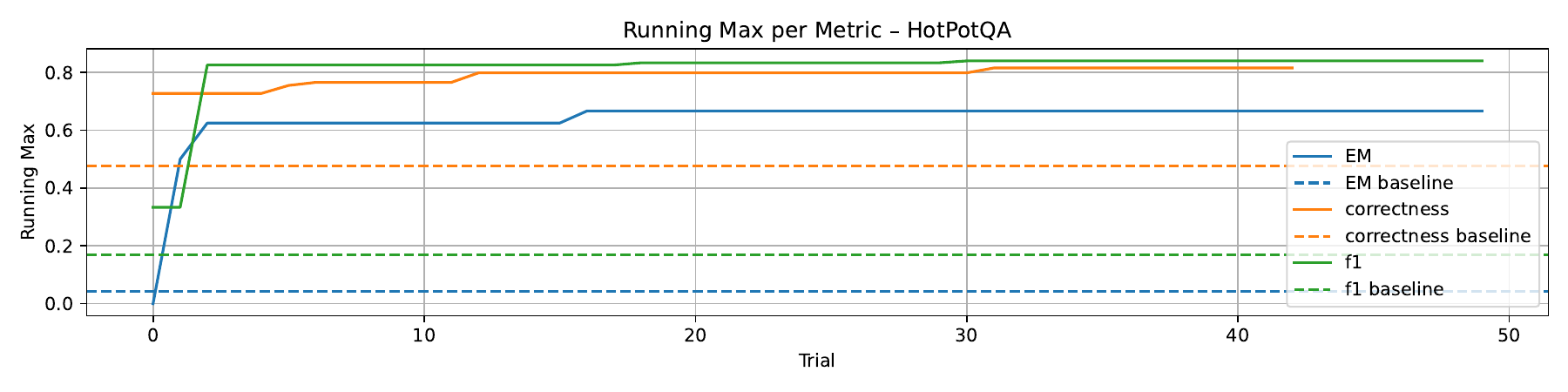}
    \caption{HotPotQA}
    \label{fig:hotpotqa}
\end{subfigure}

\caption{Running maximum performance curves for Musique, TwoWikiMultiHop, and HotPotQA.}
\label{fig:running-max-stacked}
\end{figure}

\begin{table}[h]
\centering
\begin{tabularx}{\linewidth}{@{}lXrrr@{}}
\toprule
\textbf{Benchmark} & \textbf{Metric} & \textbf{Baseline} & \textbf{Optimized} & \textbf{Relative Gain (\%)} \\
\midrule
Musique         & Correctness & 0.414 & 0.674 & 62.8 \\
Musique         & EM          & 0.000 & 0.500 & -- \\
Musique         & F1          & 0.145 & 0.654 & 351.0 \\
TwoWikiMultiHop & Correctness & 0.348 & 0.582 & 67.2 \\
TwoWikiMultiHop & EM          & 0.000 & 0.458 & -- \\
TwoWikiMultiHop & F1          & 0.148 & 0.625 & 321.6 \\
HotPotQA        & Correctness & 0.476 & 0.815 & 71.2 \\
HotPotQA        & EM          & 0.042 & 0.667 & 1496.0 \\
HotPotQA        & F1          & 0.169 & 0.840 & 396.7 \\
\bottomrule
\end{tabularx}
\caption{Training set performance of baseline and optimized configurations. Relative gain is computed as the percentage increase from baseline. Undefined where the baseline is zero.}
\label{tab:train-results}
\end{table}

Table~\ref{tab:train-results} reports performance across all benchmark–metric combinations. The baseline corresponds to the default configuration used prior to optimization. Gains are expressed as the ratio between optimized and baseline scores, where defined.

Optimization led to consistent improvements across all datasets and metrics. While the baseline settings were reasonable and manually selected, they were not tuned for the specific evaluation conditions. Relative improvements were often substantial, particularly for exact match, where several baselines were close to or exactly zero. This is largely due to a mismatch in answer style: the system’s default configuration was tuned for more conversational output, whereas the benchmarks favored shorter, drier answers. Given EM’s strictness as a metric, even factually correct responses were frequently penalized.

Despite the apparent improvements, these results should be interpreted with care. We return to this point in the discussion below.

\subsection{Hold-Out Set Performance}

\begin{table}[h]
\centering
\begin{tabularx}{\linewidth}{@{}lXrr@{}}
\toprule
\textbf{Benchmark} & \textbf{Metric} & \textbf{Train Set} & \textbf{Hold-Out Set} \\
\midrule
HotPotQA        & EM          & 0.667 & 0.583 \\
HotPotQA        & Correctness & 0.815 & 0.715 \\
HotPotQA        & F1          & 0.840 & 0.819 \\
Musique         & EM          & 0.500 & 0.375 \\
Musique         & Correctness & 0.674 & 0.596 \\
Musique         & F1          & 0.654 & 0.581 \\
TwoWikiMultiHop & EM          & 0.458 & 0.417 \\
TwoWikiMultiHop & Correctness & 0.582 & 0.482 \\
TwoWikiMultiHop & F1          & 0.625 & 0.704 \\
\bottomrule
\end{tabularx}
\caption{Performance of best configuration from each experiment on the train and held-out test sets.}
\label{tab:test-results}
\end{table}

To assess generalization, we evaluated the best configuration from each experiment on a held-out test set. Table~\ref{tab:test-results} shows the test results alongside the corresponding training scores. Gains over the baseline remained visible but were somewhat less pronounced than in training. Most metrics degraded moderately, and in one case (F1 on TwoWikiMultiHop), test performance slightly exceeded the training score. These results indicate that task-specific optimization generalizes reasonably well, even when applied to unseen examples from the same benchmark.

Some variability is likely attributable to the small size of the hold-out sets and the uneven quality of benchmark QA instances, a limitation noted throughout literature. We used a simple training setup without early stopping or regularization, which may also explain part of the observed degradation. Nevertheless, the fact that improvements persisted in most cases indicates that even basic optimization processes can yield generalizable gains. While not the primary focus of this study, these outcomes suggest that future work could explore more robust tuning regimes, especially on larger or domain-specific datasets.

A broader interpretation of these results, including implications for generalization and tuning strategies, is provided in the following discussion.

\subsection{Discussion}

As noted in Section~\ref{sec:optimization}, the optimization process used the Tree-structured Parzen Estimator (TPE), selected for its ability to navigate discrete and mixed parameter spaces. TPE was effective in identifying improved configurations, though trial-level performance was sometimes volatile. More stable or expressive optimization strategies may yield more consistent outcomes, and exploring such alternatives remains a direction for future work.

The experiments also underscored limitations in standard evaluation metrics. Exact match and F1 frequently penalized outputs that were semantically correct but phrased differently from the reference. In contrast, LLM-based correctness scores were more tolerant of lexical variation but introduced inconsistencies of their own. Several near-verbatim answers received less than full credit, suggesting that the LLM grader introduced noise, particularly around format sensitivity and implicit assumptions.

High-performing configurations often shared parameter settings, particularly for chunk size and retrieval method. However, most effects were nonlinear and task-specific, and no single configuration performed best across all benchmarks. This highlights the importance of empirical tuning in retrieval-augmented pipelines and suggests that generalization across tasks requires adaptation, not just reuse.

While full generalization remains outside the scope of this study, the results support the claim that systematic tuning is both achievable and useful in practice. The observed gains, while modest in some cases, show that configuration-level changes alone can influence downstream performance. Retrieval-augmented systems benefit from targeted, task-aware tuning, and performance–overfitting trade-offs can be managed without significant architectural change or added complexity.

\section{Conclusion}
\label{sec:conclusion}

We demonstrated that systematic hyperparameter tuning in graph-based retrieval-augmented generation systems can lead to consistent performance improvements. Cognee’s modular architecture allowed us to isolate and vary configuration parameters across graph construction, retrieval, and prompting. Applied to three multi-hop QA benchmarks, this setup enabled us to examine how tuning affected standard evaluation metrics. While improvements were observed across tasks, their magnitude varied, and gains were often sensitive to both the metric and dataset.

Looking ahead, there are several natural directions for further work. Technically, the optimization process could be extended using alternative search algorithms, broader parameter spaces, or multi-objective criteria. Our evaluation focused on well-known QA datasets, but custom benchmarks and domain-specific tasks would help probe generalization. A leaderboard or shared benchmark infrastructure for graph-augmented RAG systems could also support progress in this area.

While QA-based metrics offer a practical means of evaluating pipeline performance, they do not fully capture the complexity of graph-based systems. The variability in outcomes across configurations suggests that gains are unlikely to come from generic tuning alone. Instead, our results point to the potential of task-specific optimization strategies, particularly in settings where domain structure plays a central role. We expect that future work at the intersection of academic and applied contexts will uncover further opportunities for targeted tuning.

More broadly, we think it is useful to view this process through the lens of cognification\footnote{See Appendix~\ref{appendix:cognee}.}, a concept that describes how intelligence becomes embedded in physical systems. We see the development of frameworks like Cognee as part of a broader shift toward systems that reflect this paradigm, and their optimization plays an equally important role. The cognification of these systems will not happen through design alone, but through how they are tuned, measured, and adapted over time.
\pagebreak
\appendix
\section{Cognee}
\label{appendix:cognee}

    Cognee is an open-source Python framework for end-to-end knowledge graph (KG) construction, retrieval, and completion. Its architecture is organized around a modular Extract–Cognify–Load (ECL) pipeline. The term cognify, introduced by Kevin Kelly, describes the process of adding intelligence to already digitized systems. In Cognee, it refers to the transformation of unstructured input into structured, semantically grounded graph representations.

The Extract stage ingests heterogeneous inputs such as text, images, and audio. The Cognify stage applies schema-based transformations using Pydantic models to identify entities, relations, and attributes. The Load stage writes the resulting data to graph, relational, or vector stores. Each stage is independently configurable and replaceable, which allows adaptation to diverse data types and scaling requirements.

Cognee also supports retrieval and reasoning over constructed graphs, with integration of large language models (LLMs). Users can submit structured graph queries, prompt-based interactions, or hybrid retrievals through a single interface. The system is distributed as a Python package with containerized deployment tools and a browser-accessible user interface.

\subsection{Default Pipeline}

The default Cognee pipeline processes unstructured inputs such as text, PDFs, images, audio transcripts, and source code into structured graph representations and vector embeddings. It operates through a sequence of modular components that can be configured independently.

Inputs are ingested from local directories or remote sources and stored in a file system or object store. Metadata including file name, media type, content hash, and source location is recorded in a relational database. Files are then classified by MIME type, optionally enriched with additional metadata, and deduplicated using content hashes. They can be organized into datasets to support reuse and incremental updates.

Documents are segmented into token-limited chunks using a configurable strategy. Each chunk is processed by a language model that fills structured schema objects representing entities, relations, and summaries. These outputs are converted into graph fragments and linked to their source. Optionally, additional nodes summarizing subgraphs or lengthy input chunks can be generated and integrated into the graph.

Processing is coordinated by an orchestration layer that manages input validation, scheduling, and endpoint checks. Configuration parameters are supplied through files or API calls. Final outputs are indexed in three storage systems: a graph database for entity and relation queries, a relational store for metadata, and a vector index for similarity-based retrieval.

\begin{table}[h]
\centering
\begin{tabularx}{\linewidth}{@{}lX@{}}
\toprule
\textbf{Stage} & \textbf{Description} \\
\midrule
Ingestion & Load and normalize raw inputs into file or object storage \\
Tagging & Classify by media type, merge metadata, deduplicate, and organize into datasets \\
Chunking & Segment documents and prepare structured inputs for extraction \\
Graph Construction & Extract entities and relations using language models and assemble graph fragments \\
Indexing & Write outputs to graph, relational, and vector storage systems \\
\bottomrule
\end{tabularx}
\caption{Summary of processing stages in the default Cognee pipeline.}
\end{table}

\subsection{Retrieval Strategies}

Cognee provides several retrievers that differ in how they select and prepare context for question answering and generation. Each retriever operates over one or more indexed data sources, including vector stores and knowledge graphs. Some return retrieved content directly, while others invoke a language model to generate an answer based on the retrieved context. Multiple retrieval steps may be involved, depending on the retriever.

Retrievers can be selected through a single configuration parameter and do not require changes to the pipeline. The interface supports adjustments such as the number of retrieved items, scoring method, and prompt format where applicable. Further implementation details are available in the project repository.\footnote{\url{https://github.com/cognee-ai/cognee}} The retrievers used in the hyperparameter optimization experiments are described in Section~\ref{sec:optimization}.

\begin{table}[h]
\centering
\begin{tabularx}{\linewidth}{@{}lX@{}}
\toprule
\textbf{Retriever} & \textbf{Description} \\
\midrule
Summary-Based & Retrieves chunk-level summaries using semantic similarity \\
Chunk-Level & Retrieves original text chunks based on embedding similarity \\
Graph Neighborhood & Retrieves nodes adjacent to a matched graph entity \\
RAG & Passes retrieved text chunks to a language model for answer generation \\
Graph Completion & Retrieves graph triples and uses a language model to generate a response \\
Graph-Summary Completion & Summarizes a subgraph using a language model before generating a response \\
\bottomrule
\end{tabularx}
\caption{Retriever types supported in Cognee.}
\end{table}

\subsection{Evaluation Framework}

Cognee includes an evaluation framework for benchmarking retrieval and generation components using multi-hop question answering datasets. The framework is organized as a four-stage pipeline: corpus construction, question answering, answer evaluation, and metric aggregation. All configuration is handled through a single declarative file.

In the corpus construction stage, a dataset adapter loads question–answer pairs and their associated source texts (e.g., HotPotQA, TwoWikiMultiHop). Before each run, all memory layers are cleared and documents are reprocessed using the default pipeline. Questions and reference answers are stored in a relational table to support later evaluation.

In the answering stage, the framework instantiates a specified retriever, such as a vector-based method or graph-based strategy. Each question is paired with retrieved context, optionally passed through a language model, and the generated answer is recorded alongside its input and gold reference.

Evaluation supports two modes. One uses structured LLM-based grading (e.g., GEval) to score outputs on metrics such as correctness, exact match, F1, and contextual coverage. The other performs a direct LLM-based comparison and returns a correctness score, optionally with justification.

Aggregated results are computed using bootstrap estimates of mean performance and confidence intervals. Output includes both tabular summaries and an interactive dashboard with plots and error breakdowns.

The framework is modular and allows new tasks, retrievers, and evaluation strategies to be added with minimal changes to configuration.

\printbibliography
\end{document}